\documentclass[a4paper,conference]{IEEEtran}
\usepackage{cite}
\usepackage{balance}

\ifCLASSINFOpdf
\else
\fi
\usepackage{hyperref}
\hypersetup{
    colorlinks=true,
    linkcolor=blue,
    filecolor=magenta,      
    urlcolor=cyan,
    pdftitle={Overleaf Example},
    pdfpagemode=FullScreen,
    }
\usepackage{graphicx}
\begin{document}
%
\title{Aggregating Global Features into Local Vision Transformer}

\author{\IEEEauthorblockN{Krushi Patel$^{\dag}$, 
Andrés M. Bur$^{\ddag}$, Fengjun Li$^{\dag}$, Guanghui Wang$^{*}$}
\IEEEauthorblockA{$^\dag$ \textit{Department of Electrical Engineering and Computer Science, University of Kansas, Lawrence KS, USA, 66045}\\
$^\ddag$ Department of Otolaryngology–Head and Neck Surgery, University of Kansas, Kansas City, Kansas, USA, 66160\\
$^*$ \textit{Department of Computer Science, Ryerson University, Toronto ON, Canada, M5B 2K3}\\
\{krushi92, fli\}@ku.edu, abur@kumc.edu, wangcs@ryerson.ca
}
}

%


\maketitle

\begin{abstract}
Local Transformer-based classification models have recently achieved promising results with relatively low computational costs. However, the effect of aggregating spatial global information of local Transformer-based architecture is not clear. This work investigates the outcome of applying a global attention-based module named multi-resolution overlapped attention (MOA) in the local window-based transformer after each stage. The proposed MOA employs slightly larger and overlapped patches in the key to enable neighborhood pixel information transmission, which leads to significant performance gain. In addition, we thoroughly investigate the effect of the dimension of essential architecture components through extensive experiments and discover an optimum architecture design. Extensive experimental results CIFAR-10, CIFAR-100, and  ImageNet-1K datasets demonstrate that the proposed approach outperforms previous vision Transformers with a comparatively fewer number of parameters. The source code
and models are publicly available at: \url{https://github.com/krushi1992/MOA-transformer}
\end{abstract}


%
\IEEEpeerreviewmaketitle

\section{Introduction}

Transformer-based architecture has achieved tremendous success in the field of natural language processing (NLP) \cite{vaswani2017attention} \cite{devlin2018bert}. Inspired by the great success of transformer in the language domain, vision transformer\cite{dosovitskiy2020image} has been proposed and achieved superior performance on the ImageNet dataset.  The vision transformer splits the image into patches and feeds into the transformer,  the same way as words token in NLP, and passes through several multi-head self-attention layers of the transformer to establish the long-range dependencies. 
\begin{figure}[!t]
\centerline{\includegraphics[width=0.9\linewidth]{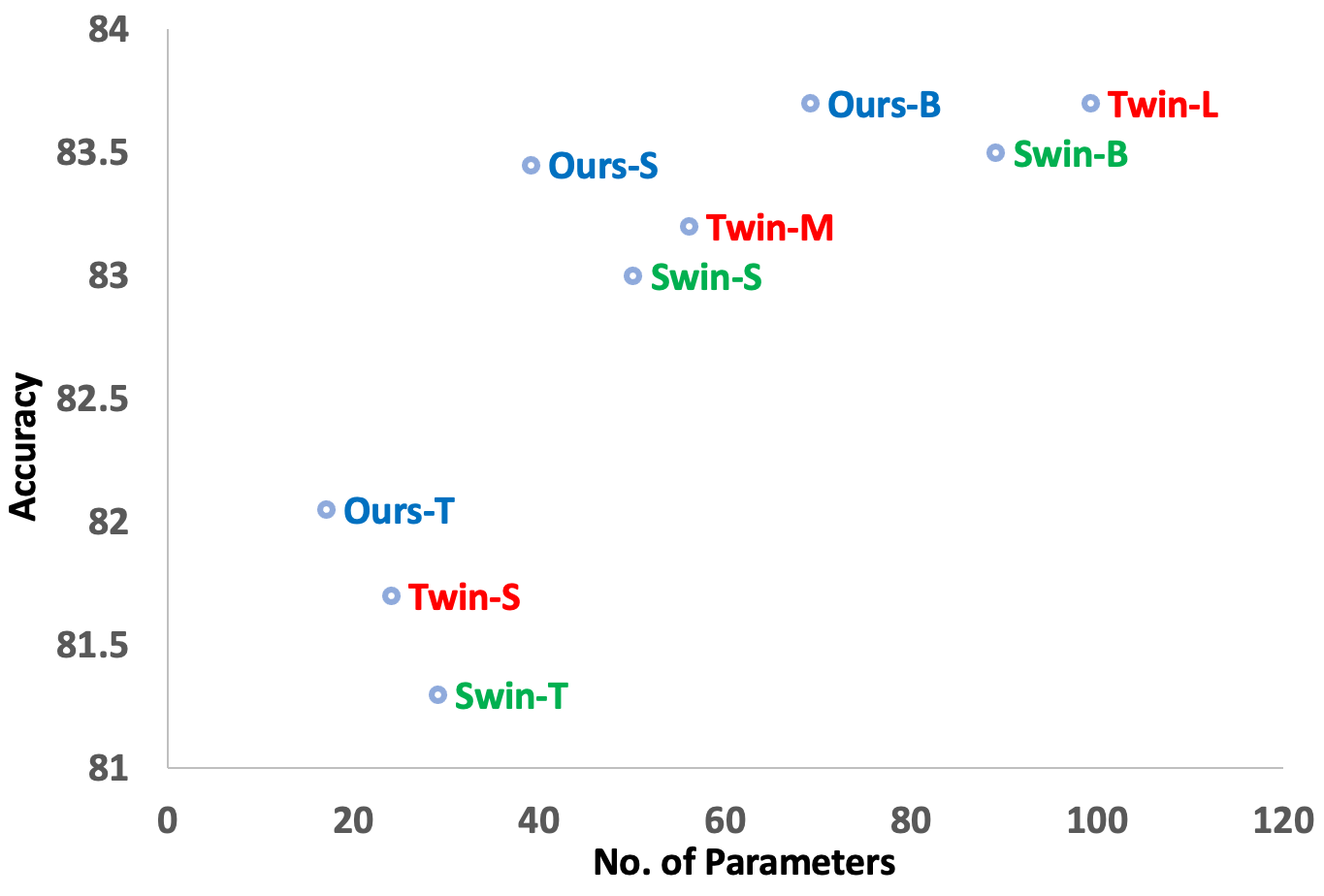}}
\caption{Graph of accuracy vs. number of parameters for various local transformer-based models. It shows that our all versions of the model: MOA-T,  MOA-S,  and MOA-B have higher accuracy and comparatively fewer number of parameters.}
\label{fig:fig1}
\end{figure}

Unlike the word token, a high-resolution image contains more pixels compared to words in the passage. This leads to an increase in the computation cost as self-attention in the transformer has quadratic complexity. To alleviate this problem, various local attention-based transformers \cite{liu2021swin} \cite{vaswani2021scaling} \cite{zhang2021nested} have been proposed with a linear computation complexity.  However,   all the proposed approaches could not establish long-range dependencies and some of them are very complicated. 

To overcome these issues in the local transformers,  we develop a very simple module, named multi-resolution overlapped attention(MOA), to generate global features. The proposed module only consists of multiplication and addition operations and is embedded after each stage in the transformer before the downsampling operation.  As the module is added only after each stage instead of each transformer layer,  it does not add much computation cost and the number of parameters.  Our experiments show that aggregating the resultant features of this module to the local transformer establish the long-range dependencies and hence significantly increases the accuracy in contrast to the total number of parameters as shown in Figure~\ref{fig:fig1}

\begin{figure*}
\centerline{\includegraphics[width=0.85\linewidth]{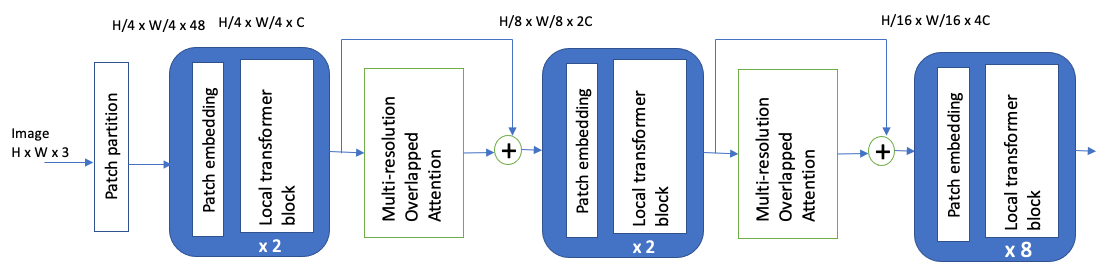}}
\caption{The architecture of the MOA-T is composed of three stages.  Each stage consists of a patch embedding/merging layer and local transformer block along with a global multi-resolution overlapped attention module after each stage except the last stage. In the beginning patch, a partition layer is included to divide the image into a fixed number of patches. }
\label{fig:fig2}
\end{figure*}
Our proposed MOA module takes the output generated by the group of local window-based attention as an input. It first converts it to a 2D feature map, and projects it to a new low-dimension feature map.  Similar to ViT \cite{dosovitskiy2020image},  the projected feature map is divided into a fixed number of patches except for a few modifications. In contrast to ViT \cite{dosovitskiy2020image},  the patch sizes of query and key-value are different.  The resolution of the patches in the query is the same as the window size used in the local transformer layer.  In contrast, the resolution of patches in key-value is slightly larger than the query patch and overlapped.  The hidden dimension of the MOA global attention module is kept the same as the previous transformer layer. Therefore, the resultant features are directly aggregated to the output of the previous transformer layer. 

Extensive experiments show that keeping the key-value patches slightly larger with overlap to each other leads to significant performance gain due to small information exchange between two neighborhood windows. In short, our method exploits the neighborhood information along with global information exchange between all non-local windows by embedding the proposed MOA mechanism in the local transformer.

The contributions of the proposed approach are summarized as below:

\begin{enumerate}

\item We propose a multi-resolution overlapped attention (MOA) module that can be plugged in after each stage in the local Transformer to promote information communication along with nearby windows and all non-local windows.

\item We thoroughly study the impact of global information in local transformer using the proposed MOA module.  

\item We investigate the effect of the dimension of essential architecture components through extensive experiments and discover the optimum architecture for image classification. 
 
\item We train the proposed model from scratch on CIFAR-10/CIFAR-100 \cite{cifar} and ImageNet-1K \cite{deng2009imagenet} datasets and achieve state-of-the-art accuracy using a local transformer.
       
\end{enumerate}

\section{Related Work}

\subsection{Convolutional Neural Networks}
After the revolutionary invention of AlexNet \cite{krizhevsky2012imagenet}, convolutional neural network (CNN) has become a standard network for all computer vision tasks, such as image classification \cite{ma2022semantic}\cite{patel2022discriminative}, object detection \cite{li2021colonoscopy}, tracking \cite{zhang2020efficient}, segmentation \cite{he2021sosd}\cite{patel2021enhanced}, counting \cite{sajid2021audio}, and image generation \cite{xu2021domain}. Various versions of CNNs have been proposed to improve the performance by making it deeper and/or broader, such as VGG network \cite{simonyan2014very},  ResNet \cite{he2016deep}, Wide-ResNet \cite{zagoruyko2016wide},  DenseNet \cite{huang2017densely}, etc.  There are also several works proposed to make it more efficient by modifying the individual convolutional layer, such as dilated convolution\cite{yu2015multi}, depth-wise separable convolution \cite{chollet2017xception},  group convolution \cite{krizhevsky2012imagenet}, etc. In our work, we employ the convolutional layer along with the transformer layer to reduce the overall dimension of the feature map.  Our experiments show that the combination of convolutions and multi-head attention increases the performance. 

\subsection{Self Attention in CNN}

Self-attention mechanisms have become ubiquitous in the field of computer vision tasks. Various works \cite{gajurel2021fine}\cite{wang2018non}\cite{cao2019gcnet}\cite{woo2018cbam}\cite{sajid2021towards}\cite{fu2019dual}\cite{zhao2018psanet}\cite{ma2021miti} have been proposed that utilize either channel-based or position based self-attention layers to augment the convolution network. Non-local network\cite{wang2018non} and PSANet\cite{zhao2018psanet} model the spatial relationship between all the pixels in the feature map and are embedded the attention module after each block in CNN, whereas SENet \cite{hu2018squeeze} establishes a channel relationship in the convolution network by squeezing the features using global average pooling.  CBAM \cite{woo2018cbam}, BAM \cite{park2018bam} and dual attention network \cite{fu2019dual} employ both channel and position based attention mechanisms separately, then combine the resultant features from both attention modules using either element-wise addition or concatenation and uses the resultant features into convolution output after each stage, whereas GCNet \cite{cao2019gcnet} combines SENet \cite{hu2018squeeze} and non-local network \cite{wang2018non} together and propose the hybrid attention mechanism that aggregates the information of both channel and spatial relationships in the same attention module. 

\subsection{Vision Transformers}

Similar to AlexNet, vision Transformer (ViT) \cite{dosovitskiy2020image} has changed the perspective of researchers towards solving computer vision problems. Since then, many vision transformer-based networks have been proposed to improve accuracy or efficiency.  The ViT needs to be pre-trained on large datasets such as JFT300M \cite{sun2017revisiting} to achieve high performance.  DeiT \cite{touvron2021training} solves this problem by student-teacher setup, substantial augmentation, and regularization techniques. To train the transformer on the mid-sized dataset like ImageNet-1K from scratch, the token-to-token vision transformer \cite{yuan2021tokens} recursively aggregate neighboring tokens (patches) into one token (patch) to reduce the number of tokens. A Cross-ViT \cite{chen2021crossvit} comes up with a dual branch approach with multi-scale patch size to produce robust image features and pyramid vision Transformer (PVT) \cite{wang2021pyramid} introduces a multi-scale-based spatial dimension design similar to FPN \cite{lin2017feature} in CNN and demonstrated good performance. Furthermore,  PVT introduced a spatial reduction in key to reduce the computation cost in multi-head attention. 

Various local attention-based transformers have been introduced to alleviate the quadratic complexity issues \cite{vaswani2021scaling}\cite{liu2021swin}\cite{zhang2021nested}.  The HaloNet \cite{vaswani2021scaling} introduces the idea of a slightly larger window of key than the query in a local attention mechanism and proves its effectiveness through various experiments. In our model, the key is also calculated using a slightly larger patch, but in the context of global attention, the idea of a larger key is different from the HaloNet. A 
swin Transformer \cite{liu2021swin} proposes a non-overlapping window-based local self-attention mechanism to avoid quadratic complexity and achieve comparable performance and aggregated nested Transformer \cite{zhang2021nested} come with the multi-scale approach with block-aggregation mechanism after each stage.

Some Transformer-based works have been proposed to utilize both local and global features \cite{han2021transformer} \cite{chu2021twins}.
A Transformer in Transformer (TNT) \cite{han2021transformer} further divides the local patches (visual sentences) into smaller patches (visual words). The MHA on visual words embedding is calculated and aggregated to the sentence embedding to establish the global relationship.  The twin Transformer \cite{chu2021twins} is quite the same as ours.  However, global attention is applied after each local Transformer layer, increasing the computation cost significantly.  In contrast, we apply it after each stage, and we have slightly larger and overlapped patches in key in multi-head attention. The proposed network efficiently utilizes global information in the local transformer and achieves higher accuracy than the above-mentioned transformer-based models. 

\begin{figure}
\centerline{\includegraphics[width=1.0\linewidth]{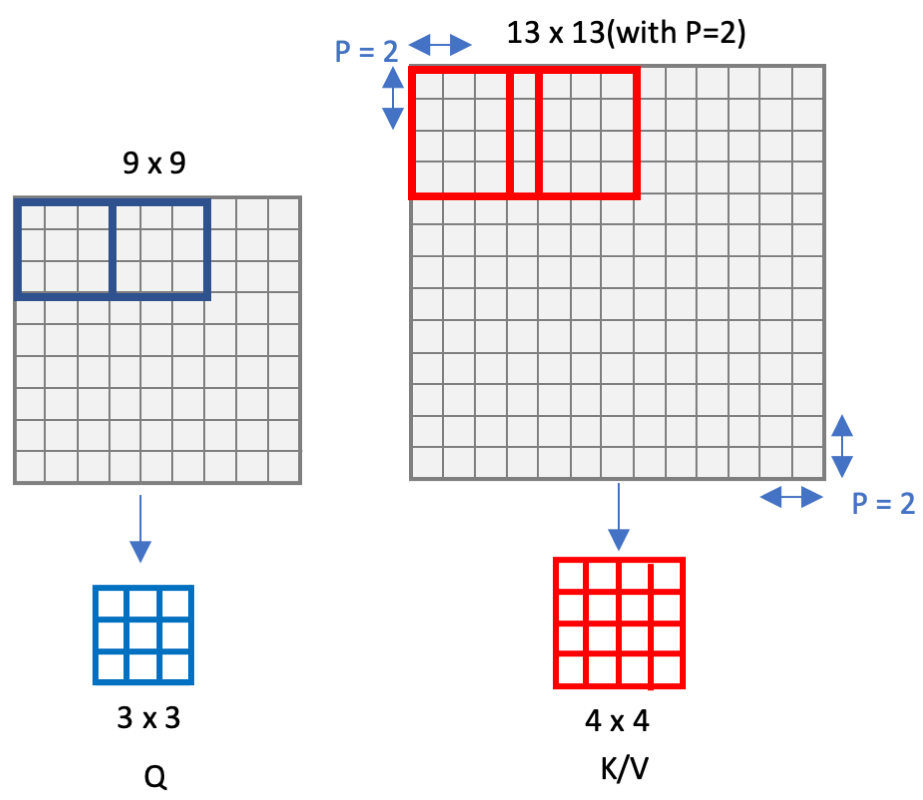}}
\caption{Patch creation for query embedding is shown in the blue,  and key/value is shown in the red for feature map size $9 \times 9$ and window size $3 \times 3$. Blue patches have the same size as the window and are non-overlapped to each other. In contrast, red patches are larger and slightly overlapped with each other. Appropriate padding is applied while creating the key-value patches.}
\label{fig:fig3}
\end{figure}

\section{Proposed Method}

We aim to provide global information exchange across all windows in the local transformer by increasing the minimal computation cost and a number of parameters.  An overview of our proposed model is shown in Figure~\ref{fig:fig2}, which shows MOA module after each stage. All stages have a similar architecture design, including patch merging layer and local transformer block except the first stage. The first stage consists of patch partition, linear embedding layer, and local transformer block. Our global MOA module is applied between each stage before the patch merging layer. 

Specifically,  the model takes an RGB image as an input and splits it into fix number of patches. Here each patch is treated as a token. In our experiment on the ImageNet dataset, we set the patch size to $4\times4$, which leads to $4\times4\times3 = 48$ feature dimensions for each patch. These row features are projected to a specific dimension C using the patch embedding layer in the first stage. The resultant features are then passed through consecutive stages consisting of patch merging layer,  local transformer block,  and MOA module in-between each stage.  Unlike swin Transformer \cite{liu2021swin}, our Transformer block employs the same self-attention mechanism as ViT \cite{dosovitskiy2020image}  without any shifted window approach.  Similar to swin Transformer, the number of tokens is reduced,  and the output dimension is doubled in the patch merging layer after each stage.  For example,  the resolution after the first,  second and third stage is $\frac{H}{2} \times \frac{W}{2}$,  $\frac{H}{4} \times \frac{W}{4}$, and $\frac{H}{8} \times \frac{W}{8}$,  respectively.  The average pooling layer is inserted at the end of the last stage,  followed by a linear layer to generate a classification score.  The detailed explanation of each element of architecture are as follows:

\subsection{Patch embedding layer}

It is a basic linear embedding layer applied to the row features of patches to project it to a specific dimension C. 

\subsection{Patch merging layer} 

Patch merging layer reduces the number of tokens by concatenating the features of $2 \times 2$ neighboring patches and doubles the number of hidden dimensions by applying a linear layer on the concatenated 4C - dimensional features.

\subsection{Local Transformer Block} 

The local transformer block consists of a local window-based standard multi-head attention module, followed by a two-layer MLP with GELU non-linearity. A layer norm is used before each multi-head attention module and each MLP with residual connection after each module.

\subsection{Multi-resolution Overlapped Attention Block}

To utilize the advantage of global information in local transformer, we apply a global attention module named multi-resolution overlapped attention (MOA) in-between each stage. The architecture of the
MOA mechanism is the same as the standard multi-head attention except for a few modifications. Similar to standard MHA, it first divides the feature map into the fixed size of patches.  However,  unlike the standard MHA, patches for generating key and value embeddings are slightly larger and overlapped,  while the patches for query embedding are non-overlapped as shown in  Figure~\ref{fig:fig3}.

As shown in the Figure~\ref{fig:fig3}, the input to MOA block is of size $W \times H \times$ hidden dim,  Where $W = \frac{W}{2}$,  $\frac{W}{4}$ or $\frac{W}{8}$, $H = \frac{H}{2}$,  $\frac{H}{4}$ or $\frac{H}{8}$, and hidden dim = 96,  192,  or 384.  Calculating query,  key,  and value embeddings directly from the input is quite expensive in computation.  For example,  in context to the ImageNet dataset,  the feature map size of the input to MOA block after the first stage is $56 \times 56 \times 96$.  Deriving query embedding directly from the input feature with a patch size 14 will lead to the resultant feature of dimension $14 \times 14 \times 96 = 18816$. Therefore,  we first reduce the hidden dimension with factor R by applying $1 \times 1$ convolution,  which reduces the computation cost. The resultant feature dimension after applying the convolution is $ H \times W \times \frac{hidden dim}{R}$.  This leads to feature size in one query patch is $14 \times 14 \times \frac {hidden dim}{R}$,  which is projected to the one-dimensional vector of size: $1 \times 1 \times hiddendim$.  The total number of the query is $\frac{H}{14} \times \frac{W}{14}$.  Similarly, the key and value vector are projected,  but the patch size is slightly larger than the query as shown in Figure~\ref{fig:fig3}.  In our model,  we set the key-value patch size to 16. Therefore,  the number of key-value will be according to the equation: $ (\frac{ H - 16 + (2 \times padding)}{stride} + 1 , \frac{ W - 16 + (2 \times padding)}{stride} + 1 )$.  Multi-head attention is applied to this query,  key,  and value embedding, followed by two-layer MLP with GELU non-linearity in between. Similar to the Transformer block, layer norm is applied along with residual connection after each MOA module.  At last, on the resultant features, $1 \times 1$ convolution is applied, followed by broadcast addition of resultant features with the output of the previous transformer block, which contains the local information.

\subsection{Relative Position Index}

We use relative position bias $B \in R ^{M^{2} \times N^{2}}$,  as used by \cite{bao2020unilmv2} \cite{hu2019local} \cite{hu2018relation} \cite{raffel2019exploring} ,  in the heads of both local and global attentions during similarity computation:

\begin{equation}
Attention(Q, K,V) = Softmax(\frac{QK^{T}}{\sqrt{d}} + B)V
\end{equation}
where $Q \in R^{M^{2} \times d}$ is a query matrix, $K, V \in R^{N^{2} \times d}$ are the key and value matrices; d is the hidden dimension, $M^{2}$ is the total number of patches in the queue and $N^{2}$ is total the number of patches in the key. 

\subsection{Architecture Detail}

By following the previous works\cite{liu2021swin}\cite{chu2021twins}, we build three versions of the model: MOA-T, MOA-S, and MOA-B for the ImageNet dataset and only two versions of the model: MOA-T and MOA-B for the CIFAR -10/100 dataset as it is quite smaller. Table ~\ref{tab: tab1} shows the architecture configurations for the CIFAR and the ImageNet datasets. In the CIFAR based models, both MOA-T and MOA-B contain the same number of Transformer layers: 12, but have a different number of hidden dimensions. In context to the ImageNet based models, the total number of layers for MOA-T and MOA-S is 12 and 24 respectively, but the hidden dimension is kept the same, whereas MOA-S and MOA-B have the same number of Transformer layers: 24, with contrast hidden dimensions 96 and 124 respectively. 

\begin{table*}
\caption{ Model configuration for CIFAR/ImageNet dataset}
\begin{center}
\begin{tabular}{ c c c c c c c c }
\hline
Model & Dataset & Input-Size & Window-Size& No. of Layers& No. of Heads&Hidden Dim& Patch -Size \\ 
\hline
MOA-T & CIFAR &$32 \times 32$ & $4 \times 4$ & [2,  2,  6,  2] & [3,  6,  12,  24]& [96,  192,  384,  768]& 1 \\
MOA-B & CIFAR & $32 \times 32$ & $4 \times 4$ & [2,  2,  6,  2] & [4,  8,  16,  32]& [128,  256,  512,  1024]& 1 \\
MOA-T & ImageNet & $224 \times 224$ & $14 \times 14$ & [2,  2,  8] & [3,  6,  12]& [96,  192,  384]& 4 \\
MOA-S & ImageNet & $224 \times 224$ & $14 \times 14$ & [2,  2,  20] & [3,  6,  12]& [96,  192,  384]& 4 \\
MOA-B & ImageNet & $224 \times 224$ & $14 \times 14$ & [2,  2,  20] & [4,  8,  16]& [128,  256,  512]& 4 \\
\hline

\label{tab: tab1}

\end{tabular}
\end{center}
\end{table*}

\section{Experimental Evaluations}
We verify our model through extensive experiments on CIFAR-10/CIFAR-100 and ImageNet-1K for image classification. We design three architecture versions: MOA-T, MOA-S, and MOA-B, for the classification tasks. 

\subsection{CIFAR-10/100 Results}
CIFAR-10 and CIFAR-100 datasets consist of 50,000 training and 10,000 test images of resolution $32 \times 32$ with the total number of classes 10 and 100, respectively.  We train the network for 300 epochs using AdamW \cite{kingma2014adam} optimizer with an initial learning rate of 0.009 and weight decay of 0.05. We utilize a cosine decay learning rate schedular along with 20 warm-up epochs. We implemented two models: MOA-T and MOA-B for the CIFAR dataset with total batch-size 128 and stochastic drop-rate 0.2 \cite{larsson2016fractalnet}. 

Table ~\ref{tab: tab3} shows the performance of our model on the CIFAR-10 and CIFAR-100 datasets. We presented only two models with the same number of layers but with different hidden dimensions for this dataset. As shown in the table, it can be seen that both models outperform all the previous Transformer-based models by a significant amount. It improves the performance by 0.59\% and 0.98\%  on CIFAR-10 and 0.56\% and 0.23\% on CIFAR-100 for the Tiny and Base models,  respectively,  compared to Swin Transformer. For the Base model, our model achieves state-of-the-art accuracy on local vision Transformer with a comparatively fewer number of parameters and GFLOPs.  The accuracy of other models is reported by training the models from scratch with the same training setting reported in the papers \cite{liu2021swin} \cite{touvron2021training} \cite{wang2021pyramid}.

\begin{table}[!htb]
\caption{Results on CIFAR - 10/100 }
\begin{center}
\begin{tabular}{ c c c c}
\hline
Model & CIFAR-100(\%) & CIFAR-10(\%)& Parameters \\ 
\hline
Deit-T&70.33&89.2&5M \\
PVT-T&72.80&91&13M \\
Swin-T& 78.07 & 94.41 & 27.5M \\
MOA-T& \textbf{78.63} &\textbf{95}& 30M  \\
\hline
DeiT-B&71.54&93&85M \\
PVT-B&70.1 &89.87&61M \\
Swin-B&78.45&94.47&86.7M \\
MOA-B&\textbf{78.68}&\textbf{95.05}& 53M \\
\hline
\label{tab: tab3}

\end{tabular}
\end{center}
\end{table}

\subsection{ImageNet Results}
ImageNet-1K dataset consists of around 1.28M training images and 50K validation images with 1000 classes. We resize all the images to the resolution $224 \times 224$ during training. We follow the same training technique, like Swin and Twin, and train the network for 300 epochs using AdamW \cite{kingma2014adam} optimizer with a cosine learning rate schedular and 20 warmup epochs. We keep the batch-size 128 for MOA-T and 64 for MOA-S and MOA-B models per GPU. We employ a total of four GPUs together during training leads to a total batch-size 512 for MOA-T and 256 for MOA-S and MOA-B models. We utilize the same augmentation technique used by \cite{liu2021swin} such as a mixture of cutmix \cite{yun2019cutmix} and mixup \cite{zhang2017mixup} and regularization technique stochastic drop rate. We set the drop rate \cite{larsson2016fractalnet} of 0.2, 0.3, and 0.5 respectively for MOA-T, MOA-S, and MOA-B.

Table ~\ref{tab: tab4} shows our model's result and a similar Transformer-based model on the ImageNet-1K classification task. Our proposed models: MOA-T, MOA-S, and MOA-B, achieve higher accuracy than most of the Transformer-based models with significant parameter reduction. MOA-T outperforms Twin-S and Swin-T by 0.34\% with around 22\% fewer parameters. Our MOA-S improves the performance by 0.5\% and  0.3\% compared to Swin-S and Twin-M respectively, even with the lower batch size during training. Our MOA-B achieves the state-of-the-art accuracy of 83.7\% on ImageNet-1K with comparatively fewer parameters with a smaller batch size than the remaining vision transformers. Our model increases the computation cost by a negligible amount, but the performance improvement and parameter reduction are highly rewardable.

\begin{table}[t]
\caption{Results on ImageNet-1K}
\begin{center}
\begin{tabular}{ c c c c}
\hline
Model & Accuracy(\%) & Parameters & GFLOPs \\ 
\hline
Deit-Small/16 & 79.9 & 22.1M & 4.6\\
CrossViT-S&81.0&26.7M&5.6\\
T2T-ViT-14&81.5&22M&5.2\\
PVT-Small&79.8&24.5M&3.8\\
TNT-T&73.9&6.1M&1.4\\
Twins-PCPVT-S&81.2&24.1M&3.8\\
Swin-T&81.3&29M&4.5\\
Twins-SVT-S&81.7&24M&2.9\\
MOA-T&\textbf{82.05}&17M&4.8\\

\hline

T2T-ViT-19&81.9&39.2M&8.9\\
PVT-Medium&81.2&44.2M&6.7\\
TNT-S&81.5&23.8M&5.2\\
Twins-PCPVT-B&82.7&43.8&6.7\\
Swin-S&83.0&50M&8.7\\
Twins-SVT-B&83.2&56M&8.6\\
MOA-S&\textbf{83.5}&39M&9.4\\

\hline

ViT-Base/16 &77.9&86.6M&17.6\\
Deit-Base/16&81.8&86.6M&17.6\\
T2T-ViT-24&82.3&64.1M&14.1\\
CrossViT-B&82.2&104.7M&21.2\\
PVT-Large&81.7&61.4M&9.8\\
TNT-B&82.9&65.6M&14.1\\
Swin-B&83.3&15.4M&83.7\\
Twins-SVT-L&\textbf{83.7}&99.2M&15.1\\
MOA-B&\textbf{83.7}&68M&16.2\\

\hline

\label{tab: tab4}

\end{tabular}
\end{center}
\end{table}

\section{Ablation Study}

In this section, we conduct ablation experiments to understand the effect of the dimension of each component, such as window size, the overlapped area between the key-value patches, and the reduction factor in global attention, in our model. We employ the Tiny model to perform all ablation experiments,  and all the experiments are performed either on CIFAR-100 or ImageNet dataset. The training configurations remain the same as reported in the experiment section. 

\subsection{Window-size}

The sequence length of the local-Transformer is one of the essential factors on which computation cost relies. As the sequence length increases, computation cost in the self-attention mechanism increases as well. In a local vision Transformer, sequence length depends on the window size.  There is always a trade-off between the accuracy and computation cost based on the sequence length. We perform experiments with various window sizes in our model and find that $4 \times 4$ and $14 \times 14 $ window size works well on CIFAR-100 and ImageNet datasets, respectively, as shown in Table ~\ref{tab: tab5}.  Furthermore, we remove the stages where the window size is greater than the feature map size to significantly reduce the number of parameters.

\begin{table}[t]
\caption{Results with different window-size on ImageNet }
\begin{center}
\begin{tabular}{ c c c c c}
\hline
Window-Size & Dataset & No. of Stage & Accuracy & Parameters \\ 
\hline
$2 \times 2$ & CIFAR -100& 4 & 76.04 & 29.7M\\
$4 \times 4$ & CIFAR-100& 4 & 78.61 & 30M \\
$8 \times 8$ & CIFAR-100& 3 & 76.02 & 16M\\
$7 \times 7$ &ImageNet& 4 & 81.4 & 31M\\
$14 \times 14$ & ImageNet&3 & 82.07 & 17M \\
$28 \times 28$ & ImageNet&2 & 78.2 & 6M\\

\hline
\label{tab: tab5}

\end{tabular}
\end{center}
\end{table}

\subsection{Overlapped Portion}

To initiate the neighborhood information transmission, we propose to use slightly larger and overlapped keys. To investigate the effect of the portion of the overlapped area, we perform experiments with different percentages of overlapped portions in keys as shown in Table ~\ref{tab: tab6}. It can be seen from the results that the performance is increased in terms of accuracy as the percentage decreases, which means only slight information exchange between the neighborhood windows are required to improve the performance.  Furthermore, fewer overlapped portions decrease the sequence length, which reduces the number of parameters and GFLOPs. 

\begin{table}[t]
\caption{Results on CIFAR-100 with different percentage of the overlapped portion}
\begin{center}
\begin{tabular}{ c c c}
\hline
\% Overlap &Accuracy & Parameters \\ 
\hline
~17\% & 78.63 & 30.05M\\
~33\% & 78.52 & 30.06M\\
~50\% & 78.38 & 30.08M\\
~66\% & 78.38 & 30.59M\\

\hline
\label{tab: tab6}

\end{tabular}
\end{center}
\end{table}

\subsection{Reduction}

Before the MOA global attention, the hidden dimension is reduced to decrease the number of parameters and computation cost. Table ~\ref{tab: tab7}  shows the performance of our model with various values of R. From the result, it is evident that R = 32 achieves the best result with comparatively less number of parameters and computation cost than a smaller value of R.
\begin{table}[t]
\caption{Results with different window-size on CIFAR-100}
\begin{center}
\begin{tabular}{ c c c}
\hline
Reduction &Accuracy & Parameters \\ 
\hline
8 & 78.38 & 31.67M\\
16 & 78.34 & 30.59M\\
32 & 78.63 & 30.06M\\
64 & 78.51& 29.78M\\
num-heads&78.41&31.43M\\

\hline
\label{tab: tab7}

\end{tabular}
\end{center}
\end{table}

\begin{table}[!t]
\caption{Significance of global attention and overlapped patches}
\begin{center}
\begin{tabular}{ c c c}
\hline
Model&Accuracy & Parameters \\ 
\hline
Without Global & 75.56 & 27M \\
With Global (ViT) & 78.34 & 30.59M\\
With Global (Ours) & 78.63 & 30.06M\\

\hline
\label{tab: tab8}

\end{tabular}
\end{center}
\end{table}

\subsection{Effect of Overlapped Key-Value}

To verify the effect of overlapped and larger key-value patches,  we train the model without overlapping patches and compare the results. Furthermore, we also conduct an experiment without applying global attention in-between each stage to verify the significance of global information exchange.  From the result in Table ~\ref{tab: tab8},  it can be seen that including global attention and overlapped key-value patches achieve the best performance.

\section{Conclusion}
The paper has investigated the effect of aggregating global information in local Transformer after each stage and neighborhood pixel information transmission. We have also proposed a multi-resolution overlapped attention (MOA) module that can be plugged in after each stage in the local transformer to promote information communication along with nearby windows. Our results show that both types of features: global and local, are crucial for image classification. As a result, exploiting both features leads to significant performance gain on the standard classification datasets such as CIFAR10/100 and the ImageNet with comparatively fewer parameters.






%



\balance
{\small
\bibliographystyle{IEEEtranS}
\bibliography{egbib}
}
\balance


\end{document}